\documentclass{article}
\usepackage{amsfonts}       
\usepackage{nicefrac}       
\usepackage{microtype}      
\usepackage{amsmath}
\usepackage{amssymb}

\usepackage[ruled,vlined]{algorithm2e}

\usepackage{graphicx}
\usepackage{color}
\usepackage{stackengine}
\usepackage{cite}

\usepackage[a4paper, total={6.5in, 10in}]{geometry}
\newcommand{\R}{\mathbb{R}}

\newcommand{\E}{\operatorname{E}}

\newcommand{\e}{\mathrm{e}}

\newcommand{\vct}[1]{\mathbf{#1}}
\newcommand{\norm}[1]{\left \|#1 \right \|}

\newcommand{\mtx}[1]{\mathbf{#1}}

\newcommand{\argmin}{\displaystyle \text{arg min}}

\newcommand{\calD}{\mathcal{D}}

\newcommand{\calN}{\mathcal{N}}
\newcommand{\calP}{\mathcal{P}}

\newcommand{\vb}{\vct{b}}

\newcommand{\vr}{\vct{r}}

\newcommand{\vx}{\vct{x}}

\newcommand{\vz}{\vct{z}}

\newcommand{\mA}{\mtx{A}}

\newcommand{\mP}{\mtx{P}}
\newcommand{\mQ}{\mtx{Q}}
\newcommand{\mR}{\mtx{R}}
\newcommand{\mS}{\mtx{S}}

\newcommand{\mU}{\mtx{U}}

\newcommand{\mW}{\mtx{W}}

\newcommand{\mY}{\mtx{Y}}

\newcommand{\mDelta}{\mtx{\Delta}}

\newcommand{\mId}{{\bf I}}

\newcommand{\mzero}{{\bf 0}}

 \newtheorem{lemma}{Lemma}
 \newtheorem{theorem}{Theorem}

 \def \endprf{\hfill {\vrule height6pt width6pt depth0pt}\medskip}

\newcommand{\sd}{\mathbf{sd}}
\newcommand{\sr}{\mathbf{sr}}
\newcommand{\mj}{M_j}
\newcommand{\wm}{\widetilde{M}}
\newcommand{\wn}{\widetilde{N}}

\newcommand{\bfcalP}{\mathbf{\calP}}
\newcommand{\dbd}{\begin{bmatrix}
\mS_1 & 0 & \cdots & 0\\
0 & \mS_2 & \cdots & 0\\
\vdots & \vdots & \ddots & \vdots \\
0 & 0 & \cdots & \mS_J
\end{bmatrix}}

\DeclareMathOperator{\rank}{rank}
\DeclareMathOperator{\vecd}{vec}

\title{Localized sketching for matrix multiplication and  ridge regression}
\author{
Rakshith S Srinivasa \thanks{This work was supported in part NSF  CCF-1718771, NSF DMS 18-00872 and in part by C-BRIC, one of six centers in JUMP, a Semiconductor Research Corporation (SRC) program sponsored by DARPA. The authors would also like to acknowledge Agniva Chowdhury for their valuable discussion and feedback during the preparation of this paper.}, Mark A Davenport and Justin Romberg 
}
\renewcommand\footnotemark{}
\date{}

\begin{document}
\maketitle

%

%

\begin{abstract}
 We consider sketched approximate matrix multiplication and ridge regression in the novel setting of localized sketching, where at any given point, only part of the data matrix is available. This corresponds to a block diagonal structure on the sketching matrix. We show that, under mild conditions, block diagonal sketching matrices require only $O(\sr / \epsilon^2)$ and $O(\sd_{\lambda}/\epsilon)$ total sample complexity for matrix multiplication and ridge regression, respectively. This matches the state-of-the-art bounds that are obtained using global sketching matrices. The localized nature of sketching considered allows for different parts of the data matrix to be sketched independently and hence is more amenable to computation in distributed and streaming settings and results in a smaller memory and computational footprint. 
\end{abstract}

\section{Introduction}

Efficient linear algebraic computations are of fundamental importance in machine learning and signal processing applications. This has led to a rise in randomized linear algebraic methods that aim to solve large problems only approximately, but with much less time complexity compared to standard methods (see \cite{Woodruff:2014:STN:2693651.2693652, wang2017sketched,7355313,pmlr-v80-chowdhury18a} and references therein). In this work, we consider two specific examples: sketched matrix multiplication \cite{optimal} and ridge regression \cite{sharper} but with additional constraints on the sketching matrices that arise in the context of distributed data acquisition. Formally, if $\mW \in \R^{\wn \times m}$ and $\mY \in \R^{\wn \times p}$, computing the product $\mW^T\mY$ takes $O(mp\wn)$ time, which can be prohibitive for large $\wn$. The sketched version then aims to find matrices $\mS \in \R^{\wm \times \wn}$ such that 
\begin{equation}
\label{eq:app_mtx_product}
    \norm{(\mS\mW)^T (\mS\mY) - \mW^T\mY} \leq \epsilon \norm{\mW}\norm{\mY}.
\end{equation}
Computing the sketched matrix product $(\mS\mW)^T (\mS\mY)$ then takes only $O(mp\wm)$ time (not accounting the time to compute $\mS \mW$ and $\mS \mY$ themselves). State-of-the-art bounds show that $\wm = O(\max( \sr(\mW), \sr(\mY))/\epsilon^2)$ suffices, where $\sr(\cdot)$ is the stable rank of a matrix (defined in Section \ref{sec:main} and is a stable alternative for the rank). Similarly, given $\mA \in \R^{\wn \times d}$ with $\wn \gg d$ and $\vb \in \R^{\wn}$, the ridge regression problem is 
\begin{equation}
    \label{eq:ridge_reg_def}
    \vx_* = \underset{\vx \in \R^d}{\argmin} ~f(\vx) := \norm{\mA \vx - \vb}^2 + \lambda \norm{\vx}^2
\end{equation} and can be solved in $O(\wn d^2)$ time. The sketched problem instead seeks to find matrices $\mS \in \R^{\wm \times \wn}$ such that solving 
\begin{equation}
    \label{eq:sketch_ridge_reg_def}
    \Hat{\vx} = \underset{\vx \in \R^d}{\argmin} ~f_{\mS}(\vx) := \norm{\mS \mA \vx - \mS \vb}^2 + \lambda \norm{\vx}^2
\end{equation} yields
\begin{equation}
    \label{eq:goal}
    f(\Hat{\vx}) \leq (1 + \epsilon) f(\vx_{*}).
\end{equation} The state-of-the-art bounds show that for small $\epsilon$, $\wm = O(\sd_{\lambda}/\epsilon)$ suffices, where $\sd_{\lambda}$ is the statistical dimension and is again a more stable alternative to the rank of $\mA$, as defined in Section \ref{sec:main}.

With this background in place, let us consider a scenario where the data matrix $\mA$ is naturally divided into $J$ blocks that are not all available at a single location. Let each block then be of size $N \times d$, where $\tilde{N} = JN$. Such partitioning of data into different blocks occurs naturally in many applications. For example, dynamic systems produce data that evolve over time. To store the entire data before sketching it would require large amounts of memory [9]. It would be of use to sketch the system as it evolves, leading to a natural partition. In yet another application, consider the square kilometer array [11]. This array consists of antennas distributed across the continents of Australia and Africa. To handle the massive data rates ($157$ TB/s), it is desirable to sketch the data locally at each antenna and then transmit to the central processing location. In distributed systems that use edge-cloud architecture, edge nodes collect data that needs to be communicated to the cloud for inference. The communication requirements can be made smaller if the data at each edge node is compressed to an “optimal” dimension. 

\begin{figure*}
    \centering
    \includegraphics[scale=0.4]{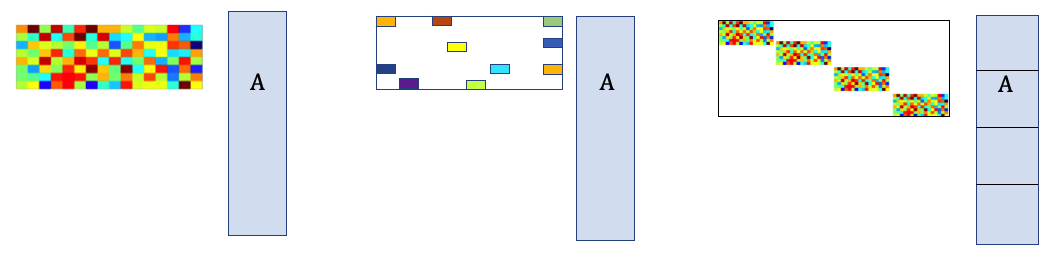}
    \caption{Existing sketching strategies such as dense sub-Gaussian, SRHT matrices (left) and sparse sketching matrices (center) assume access to all or a few arbitrarily placed rows of $\mA$. However, our localized model (right) needs access to only well-separated parts of the data matrix.}
    \label{fig:blocking}
\end{figure*}

A feature of existing sketching methods (including those that use fast Johnson-Lindenstrauss matrices such as Subsampled Randomized Hadamard Transform (SRHT) \cite{ailon2006approximate} and sparse sketching matrices \cite{clarkson2017low}) is that they need access to all or an arbitrary subset of the rows of $\mA$ (See Figure \ref{fig:blocking}). Clearly, this is unsuitable for an application with distributed data.  This leads us to ask the following questions: Is there a way to adapt sketching techniques to such applications? What is the best way to model dimensionality reduction for such applications? Two na\"{i}ve ways are readily available: i) Since each block is of size $N\times d$, its rank is upper bounded by $d$. One could obtain a subspace embedding for each block and communicate these sketched blocks to the central node. The resulting dimension of the aggregated data is then $O(Jd/\epsilon^2)$, since each block needs to be sketched to $O(d/\epsilon^2)$,   ii) Sketch each data block separately, and \textbf{add} the resulting sketches at the central node instead of aggregating them. In fact, this results in a sketch of the entire data matrix $\mA$. Using existing bounds, one can conclude that the final sketch needs to be $O(d/\epsilon^2)$, which again requires each data block also to be sketched to $O(d/\epsilon^2)$. 

A major drawback of both of the above approaches is that they do not take advantage of the inherent low dimensionality of the entire matrix $\mA$, resulting in a sketch size of $O(d/\epsilon^2)$ for each data block. Our observation is that it should be possible to lose information locally, while still retaining all the information about $\mA$ globally. This is exactly what we address in this paper: we show theoretically that it is possible for the each of the blocks to be sketched to $O(d/J\epsilon^2)$. This implies that the sketch obtained from a single block may not be big enough to provide a subspace embedding for that block. Yet, an embedding of the entire matrix $\mA$ can be obtained, once the sketches from the individual blocks are aggregated. Hence, our work aims to initiate a study of how to extend sketching methods to distributed data acquisition scenarios.

Our proposal is to impose a block diagonal structure on the sketching matrix $\mS$. We  denote such a sketching matrix as $\mS_D$. We then partition the data matrices $\mW$, $\mY$ and $\mA$ analogously. This results in sketches of the form
\setlength{\abovedisplayskip}{5pt}
\setlength{\belowdisplayskip}{5pt}%
\begin{align}
\label{eq:dbd_model}
    \mS_D \mA = \dbd \begin{bmatrix} \mA_1 \\ \mA_2 \\ \vdots \\ \mA_J \end{bmatrix} =  \begin{bmatrix} \mS_1 \mA_1 \\ \mS_2\mA_2 \\ \vdots \\ \mS_J \mA_J \end{bmatrix}.
\end{align}
We assume that $\mA_j \in \R^{N \times d}$ where $\wn = JN$ and $\mS_j \in \R^{M_j \times N}$ such that $\sum_j M_j = \wm$, although our results extend to the case where the $\mA_j$'s are of different sizes. Further, in our paper we assume that the non-zero entries of the matrix $\mS_D$ are drawn from the Gaussian distribution. Our goal is to study the sample complexities $\wm_j$ required to achieve similar guarantees as those in \cite{optimal} and \cite{sharper} for dense (non-block diagonal) sketching matrices.

Apart from the structural advantages described above, computing the product $\mS_D\mA$ can also be much cheaper when compared to an unstructured random projection. For generic $\mS_j$, the sketch $\mS_D \mA$ can be computed  in time $O(Nd\wm )$, as compared to the $O( \wn d \wm)$ required for a dense, unstructured sketch.  Second, the computation is trivial to parallelize into $J$ blocks, each requiring $O(NdM_j)$ time.  For large problems with low effective rank, when we can take $M_j=O(\log N)$, this gives us a sketch with structured randomness competitive with methods that use SRHT and sparse embedding matrices \cite{Woodruff:2014:STN:2693651.2693652}. Furthermore, the blocks themselves could be designed to be fast transforms. Owing to these computational advantages, blocking could be a strategy by itself. 
\subsection{Related work}

There is a vast and growing literature on sketching techniques. Here we briefly review some of the work most relevant to ours in the context of our setting. Note that while sketching can also be used as a pre-conditioning method~\cite{7355313}, here we will only address ``sketch and solve'' methods where the original problem is (approximately) solved in a reduced dimension. 

Sketching methods for solving ordinary least squares problems are well summarized in \cite{Woodruff:2014:STN:2693651.2693652}. However, as noted in \cite{sharper}, solutions for sketched ridge regression problems are more relevant in practice since regularization is often necessary. Similar to \cite{sharper}, we address this problem but in the setting where the sketching matrix is block diagonal. We provide conditions on the matrix $\begin{bmatrix}\mA & \vb \end{bmatrix}$ under which such structured matrices can have the same sample complexity as \cite{sharper}.

Our work is closely related to that of \cite{RIPBD} which studies the restricted isometry property (RIP) of block diagonal matrices. These results can be used to directly obtain subspace embedding guarantees for block diagonal matrices. However, this approach requires a sample complexity dependent on the rank of $\mA$ and not its approximate rank. For large matrices with fast spectral decay, this dependency can lead to sub-optimal sample complexity. Another difference is that we consider block diagonal matrices that have different sized blocks, while \cite{RIPBD} assumes that all the blocks are of the same size. One of the main conclusions of our paper is that choosing the block sizes in a data dependent fashion leads to improved (optimal) sample complexity. 


A statistical analysis of sketched ridge regression in a distributed setting is provided in \cite{wang2017sketched}. This work considers the ridge regression problem in the multivariate setting (where $\vb$ and $\vx$ are matrices) and analyzes model averaging in the case of distributed computation of the sketched ridge regression solution. In this setting, various processors each solve the problem with a part of the data and the estimators are then communicated to a central agent. In contrast, we consider a scenario where the estimate is computed by the central agent with only sketched data sent from various nodes.  

Another work that is similar in spirit to ours and addresses sketched regression in a distributed setting is \cite{mcwilliams2014loco}. The setting considered in this work lies somewhere between that of \cite{wang2017sketched} and ours. It considers multiple processors solving the ridge regression problem with different parts of the data similar to \cite{wang2017sketched}, but also assumes that the data used by each processor is available to all other processors in a sketched form. In contrast, in our work, the sketched data from all the nodes is available to only a central computing agent.

A complimentary line of work focuses on the same problem but where $\wn \ll d$. In \cite{Chen:2015:FRA:3020847.3020869}, a sketching based algorithm is proposed that achieves a relative error guarantee for the solution vector. This result is further improved in \cite{pmlr-v80-chowdhury18a}. Sketching has also been applied in the context of kernel ridge regression, where the data points are mapped to higher dimensional feature space before solving the regression problem. Sketching is used to reduce the number of such high dimensional features in  \cite{krr_drineas} and \cite{doi:10.1137/16M1105396}. Sampling and rescaling of features is considered in  \cite{krr_drineas}. Random feature maps are also used to construct pre-conditioners in \cite{doi:10.1137/16M1105396} to solve kernel ridge regression, where it is shown that a number of random feature maps proportional to the effective rank of the kernel matrix suffices to obtain a high quality pre-conditioner. While our work targets a different setting (where $\wn \gg d$) and requires a different set of analytical tools, it is noteworthy that our guarantees involve a similar dependence on the stable rank of the underlying data matrix. 
\vspace{-2pt}
\section{Main results}
\label{sec:main}
\vspace{-2pt}
Our main contribution is theoretical analysis of the block model described in \eqref{eq:dbd_model}. A na\"{i}ve strategy to analyze block diagonal matrices is to treat each block $\mA_j$ separately and use a number of random projections proportional to its effective rank.  But this would not take advantage of the low dimensional structure of the full matrix $\mA$, resulting in a highly suboptimal sample complexity. Instead,  we show that under mild assumptions on $\mA$, the total sample complexity of $\wm$ of the matrix $\mS_D$ can match the existing bounds mentioned above. 

\subsection{Stable rank, statistical dimension and incoherence}

Before we can state our main results, we need to define a few quantities that characterize the \textit{complexity} of matrix multiplication and ridge regression problems.

\textbf{Stable rank of a matrix:} The stable rank of a matrix $\mW$ is defined as $
    \sr(\mW) = \frac{\norm{\mW}_F^2}{\norm{\mW}^2}
$. Note that $\sr(\mW) \leq \rank(\mW)$. For matrices with a flat spectrum, the stable rank equals the rank of the matrix. However, if the singular values decay, then the stable rank captures the effective low dimensionality of the matrix, even when it is technically full rank. 

\textbf{Statistical dimension of the ridge regression problem:} The ridge regression problem defined in \eqref{eq:ridge_reg_def} can be reformulated as 
\begin{equation*}
    \label{eq:reform_OLS}
    \underset{\vx \in \R^d}{\min}~\norm{\begin{bmatrix}\mA \\ \sqrt{\lambda}\mId_d \end{bmatrix}\vx - \begin{bmatrix}\vb \\ \mzero \end{bmatrix}}^2 \Leftrightarrow \underset{\vx \in \R^d}{\min}~ \norm{\widetilde{\mA}\vx - \widetilde{\vb}}^2.
\end{equation*} 
 The scalar multiple of the identity on the bottom of $\widetilde{\mA}$ means it will technically be rank $d$.  But in some sense, a more nuanced notion of rank would count dimensions in the column space of $\widetilde{\mA}$ that have singular values greater than $\sqrt{\lambda}$ differently that those with singular values less than $\sqrt{\lambda}$.  One way to make to bring this distinction out is through the \textit{statistical dimension}
\begin{equation*}\sd_\lambda = \sum_i \frac{\sigma_i^2}{\sigma_i^2 + \lambda}. \end{equation*}
In the sum above, if $\sigma_i^2\gg\lambda$, then the contribution for that term is approximately one, while if $\sigma^2\ll\lambda$, it is essentially zero.  This allows us to interpret $\sd_\lambda$ as a kind of ``effective rank''.  Note that $\sd_\lambda \leq \rank(\mA)$ and can be much lower than $\rank(\mA)$. While making $\lambda$ very large can of course make $\sd_\lambda$ very small, this also introduces a larger bias in the estimates provided by \eqref{eq:ridge_reg_def} and \eqref{eq:sketch_ridge_reg_def}, driving both of their solutions to zero.  Choosing the $\lambda$ that balances this bias-variance trade-off is equally important in sketched and non-sketched ridge regression.

\textbf{Incoherence of the data matrices:}
 In randomized sampling schemes, the sampling probability of each row depends on the corresponding \textit{leverage score}, which is the $\ell_2$ norm of the corresponding row of an orthobasis $\mU$ for $\mA$. Leverage scores highlight the relative importance of each row of $\mA$.

Block diagonal matrices can be thought of as a generalization of sampling matrices. Instead of a single row, each block now accesses a submatrix of $\mA$. Instead of using uniformly sized diagonal blocks $\mS_j$, we show that a relative importance term associated with each block $\mA_j$ similar to leverage scores dictates the number of random projections $\mj$ required to attain optimal sample complexity. Let $\mU$ be an orthobasis for the column space of the matrix $\mA$. Let $\mU = [\mU_1^T \ \mU_2^T \ \cdots \ \mU_J^T ]^T$, where $\mU_j \in \R^{N\times d}$. We will show that the corresponding relative importance parameter, which we term as \textit{coherence} of $\mU_j$, is  \begin{equation*}
    \Gamma(\mU_j) = \min \left ( \norm{\mU_j}_{\infty}^2N,  \norm{\mU_j}_2^2 \right ).
\end{equation*}
Here, $\norm{\mU_j}_{\infty}$ denotes the element-wise infinity norm and $\norm{\mU_j}_2$ denotes the spectral norm. We can observe that 
\begin{equation} \frac{1}{J} \leq \max_j \Gamma(\mU_j) \leq 1. \end{equation} When the  $\Gamma(\mU_j)$'s are all close to $1/J$, the columns of $\mU$ are incoherent, or not too aligned with respect to the standard basis vectors. On the contrary, when they are close to $1$, then there are vectors in the column space of $\mU$ which are close (in an inner product sense) to the standard basis vectors. We describe bases $\mU$ that have small coherence parameters as being \textit{incoherent}. We will show that as long as the coherence is not too high, the sample complexity of block diagonal matrices can match that of generic sketching matrices. 

\textbf{Number of random projections:}
Low values of the coherence parameter (highly incoherent bases) indicate relative uniformity in the importance of the blocks. For such subspaces, it would be reasonable to expect that roughly the same number of random projections can be drawn from each data block $\mA_j$. On the other hand, when the coherence parameters $\Gamma(\mU_j)$ have a high dynamic range, it can be expected that the number of random projections from each block should be proportional to the corresponding $\Gamma(\mU_j)$. This is precisely our proposed strategy to design the number of random projections $\mj$. We propose that $\mj$ can be chosen as
\begin{equation}
\label{eq:strat}
    \mj = M_0 \Gamma(\mU_j)
\end{equation} for some constant $M_0$ that we will determine later. Our theoretical results state that block diagonal sketching matrices can achieve optimal sample complexity when $\mj$'s are designed as in \eqref{eq:strat}. This is also reminiscent of sampling algorithms, where the sampling probability of each row is proportional to the corresponding leverage score.

\subsection{Sample complexity bounds for localized sketching}
\label{subsec:bounds}

\subsubsection*{Localized sketching for matrix multiplication}

 Some of the earlier works that addressed this problem required $\mS$ to be of size $\Omega \left ( \frac{r(\mW) + r(\mY)}{\epsilon^2} \right ) \times \wn$ where $r(\cdot)$ denotes the rank of the matrix. However, matrices with high ranks can still be approximately low dimensional, as indicated by their stable rank. In \cite{optimal} it is shown that the sample complexity of $\mS$ in \eqref{eq:app_mtx_product} (under certain distributions) depends only on the \textit{stable ranks} of the matrices. They describe distributions $\calD$ that satisfy 
\begin{equation}
    \label{eq:stbl_rnk_mtx_prduct}
    \underset{\mS \sim \calD}{\mathbb{P}} \bigg( \norm{(\mS \mW)^T (\mS \mY) -  \mW^T\mY} > \epsilon \norm{\mW}\norm{\mY} \nonumber  \sqrt{(1 + \sr(\mW)/k)} \sqrt{(1 + \sr(\mY)/k)} \bigg )  < \delta
\end{equation}  for any desired $k$ and a suitable $\wm$.  When $\mS$ is a dense matrix with sub-Gaussian entries, this holds for $\wm = \Omega(\frac{k + \log(1/\delta)}{\epsilon^2})$. Then, for $k = \max(\sr(\mW), \  \sr(\mY) )$, $\mS$ satisfies \eqref{eq:app_mtx_product}. Hence, to achieve a relative error in the spectral norm, $\mS$ only needs to have a number of rows proportional to the stable ranks of $\mW$ and $\mY$.

Our first main result is such a guarantee for block diagonal sketching matrices. Unlike the distributions proposed in \cite{optimal}, block diagonal distributions cannot be both oblivious to the data matrices and have optimal sample complexity. A na\"{i}ve way to achieve \eqref{eq:stbl_rnk_mtx_prduct} when $\mS$ is block diagonal is to use triangle inequality:
\begin{align*}
    \Big\| (\mS_D \mW)^T & (\mS_D \mY) -  \mW^T\mY \Big\|  \leq  \sum_j \norm{(\mS_j \mW_j)^T (\mS_j \mY_j) -  \mW_j^T\mY_j} 
\end{align*}  
where $\mW_j$ and $\mY_j$ are corresponding blocks as in \eqref{eq:dbd_model}. However, this requires that $\mj = \Omega\left(\frac{\sr(\mW_j) + \sr(\mY_j)}{\epsilon^2}\right)$ for each $j$. This can lead to suboptimal sample complexities, as $\sr(\mW_j)$ and $\sr(\mY_j)$ can be as high as $\sr(\mW)$ and $\sr(\mY)$ themselves. We show in our analysis that we can in fact achieve 
\[ 
\wm = \sum_j \mj = \Omega \left(\frac{\sr(\mW) + \sr(\mY)}{ \epsilon^2}\right) 
\] for incoherent matrices.
 With $\mj$ designed as in \eqref{eq:strat}, we have the following result for computing approximate matrix products:
 
\begin{theorem}
\label{thm:DBD_apprx_mtx_prdct}
Fix matrices $\mW$ and $\mY$ and let $\mS_D$ be a block diagonal matrix as in \eqref{eq:dbd_model} with the entries of $\mS_j$ are drawn from the distribution $\calN(0, 1/\mj)$. Let $\mU$ be an orthobasis for the matrix $[ \mW \ \mY ]$ and $\Gamma(\mU_j)$ be the corresponding incoherence terms.  Then the tail bound \eqref{eq:stbl_rnk_mtx_prduct} holds with $\mS=\mS_D$ when $\mj$ are taken as in \eqref{eq:strat} with 
\begin{equation}
    M_0 = \Omega \left ( \frac{k \log(2/\delta)}{\epsilon^2} \right ).
\end{equation}
\end{theorem}

 We can examine the total sample complexity of $\mS_D$. Consider a highly incoherent basis $\mU$: each entry of such a basis is bounded away from $1$. Examples of such bases include orthobases of matrices with entries drawn from the Gaussian distribution and any subset of the Fourier basis. Since each column of $\mU$ has an $\ell_2$-norm of $1$, for such bases, $\norm{\mU_j}_\infty \approx 1/\sqrt{\wn}$. Then we have  $\mj \approx \frac{M_0}{J}$ and $\wm = \Omega \left ( \frac{\max(\sr(\mW),\sr(\mY)) \log(2/\delta)}{\epsilon^2} \right )$. We see that even though $\mS_D$ has a block diagonal structure, it can still have an optimal sample complexity.

\subsubsection*{Block diagonal sketching of ridge regression}

Let us now consider the sketched ridge regression problem shown in \eqref{eq:sketch_ridge_reg_def}. Let $\mU_1 \in \R^{\wm \times d}$ comprise the first $n$ rows of an orthobasis for the matrix $[ \begin{smallmatrix} \mA \\ \sqrt{\lambda}\mId_d \end{smallmatrix} ]$. Then, \eqref{eq:goal} holds with constant probability, if $\mS$ satisfies the following two conditions:
\begin{align}
\label{eq:SRR1}
    \norm{\mU_1^T\mS^T\mS\mU_1 - \mU_1^T\mU_1} & \leq \frac14, \\
\label{eq:SRR2}    
    \norm{\mU_1^T\mS^T\mS \vr^* - \mU_1^T\vr^*} & \leq \sqrt{\frac{\epsilon f(\vx^*)}{2}},
\end{align} where $\vr_* = \vb - \mA\vx^*$ and we recall that $f(\vx^*) = \norm{\mA \vx^* - \vb}^2 + \lambda \norm{\vx^*}^2$.
These conditions are well known in the randomized linear algebra community. (See \cite{sharper} Lemma 9.) Both of the above conditions on $\mS$ can be re-expressed as approximate matrix product guarantees by choosing the pair of matrices as $\mW = \mY = \mU_1$ for \eqref{eq:SRR1} and $\mW=\mU_1$ and $\mY = (\vb - \mA\vx^*)$ for \eqref{eq:SRR2}.  We now state our main result for block diagonal sketching of ridge regression problems. Let $\mA$ and $\vb$ be as defined above and let $\mU$ be an orthobasis for a basis for the range of $\left [ \mA \ \vb \right ]$ of size at most $\wn \times (d+1)$ with $\Gamma(\mU_j)$'s being the corresponding incoherence terms. 

\begin{theorem}
\label{thm:DBD}
 Let $\mU$ be an orthobasis for the matrix $[ \mA \ \vb ]$ and $\Gamma(\mU_j)$ be the corresponding incoherence terms.  Let $\mS_D$ be a block diagonal matrix as in \eqref{eq:dbd_model} with the entries of $\mS_j$ are drawn from the distribution $\calN(0, 1/\mj)$.  Let $\vx_*$ be the solution to \eqref{eq:ridge_reg_def}, and $\Hat{\vx}$ be the solution to \eqref{eq:sketch_ridge_reg_def}.  Then
\[
    f(\Hat{\vx}) \leq (1+\epsilon) f(\vx_*),
\]
with constant probability when $\mj$ obeys \eqref{eq:strat} with  $M_0 = \Omega \left ( \frac{\sd_\lambda}{\epsilon} \right )$.
\end{theorem}
As before, if $\mA$ and $\vb$ are such that the basis $\mU$ is incoherent, then the total sample complexity $\wm = \sum_j M_j = O(\frac{\sd_{\lambda}}{\epsilon})$. We are hence able to establish that though highly structured, block diagonal random matrices can in fact have optimal sample complexities.

\subsubsection*{Estimating the incoherence terms } 
\begin{figure}[ht]
    \centering
    \includegraphics[scale=0.2]{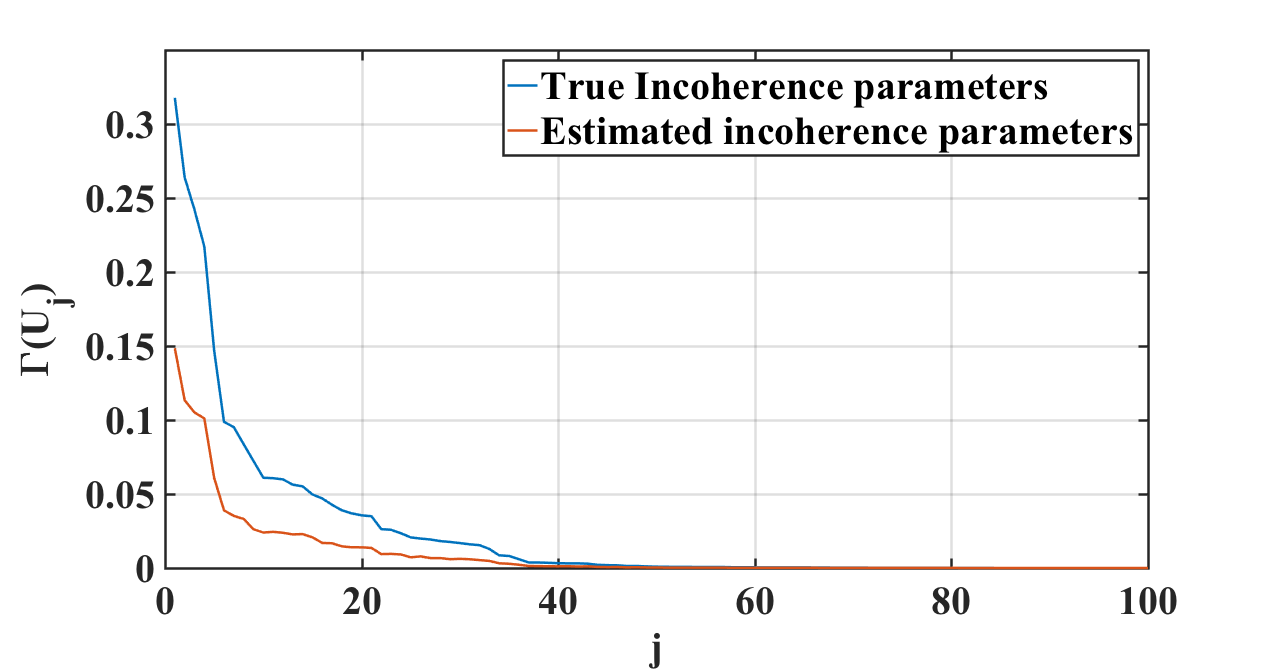}
    \caption{For a test matrix with $J = 100$, $\wn = 10000$, the true incoherence values and the estimated values are within a constant factor of each other, shown here in a sorted. Choosing the block sizes $M_j$ proportional to the estimated coherence parameters results in optimal sample complexities.}
    \label{fig:est_gamma}
\end{figure}

An important question is about how the coherence parameters $\Gamma(\mU_j)$'s can be estimated. Note that the main challenge is in computing an orthobasis for the data matrix $\mA$. We develop an algorithm to empirically estimate the $\Gamma(\mU_j)$'s to within a constant factor of the true values using a sketching based algorithm. The algorithm uses $O(d)$ fast localized random projections of the blocks $\mA_j$'s and computes an estimate of the QR factorization of $\mA$ at a central processing unit. Using the approximate R factor, the blocks $\mU_j$'s are estimated locally. The algorithm is detailed in the appendix and has a worst case time complexity of $O(\wn d \log N)$. Note that this is less than the sketch compute time $O(\wn d \wm / J)$ for $N$ not too large. In Figure \ref{fig:est_gamma}, we show the estimated incoherence parameters and the true parameters for a test matrix with $J = 100$, $\wn = 10000$. We can see that the estimated values are within a constant factor of the true $\Gamma(\mU_j)$'s. 
An important note here is that in many applications, an estimate of the $\Gamma(\mU_j)$'s may be obtained using a priori domain knowledge. Yet another insight is that if distributional assumptions on the data can be made, as common in machine learning, then $\Gamma(\mU_j)$'s can be very reliably estimated a priori \cite{RIPBD}. Any such prior information will lead to better sample complexities as compared to the na\"{i}ve techniques described in the introduction.

\section{Experiments}
\begin{figure}[t]
    \centering
    \includegraphics[scale=0.15]{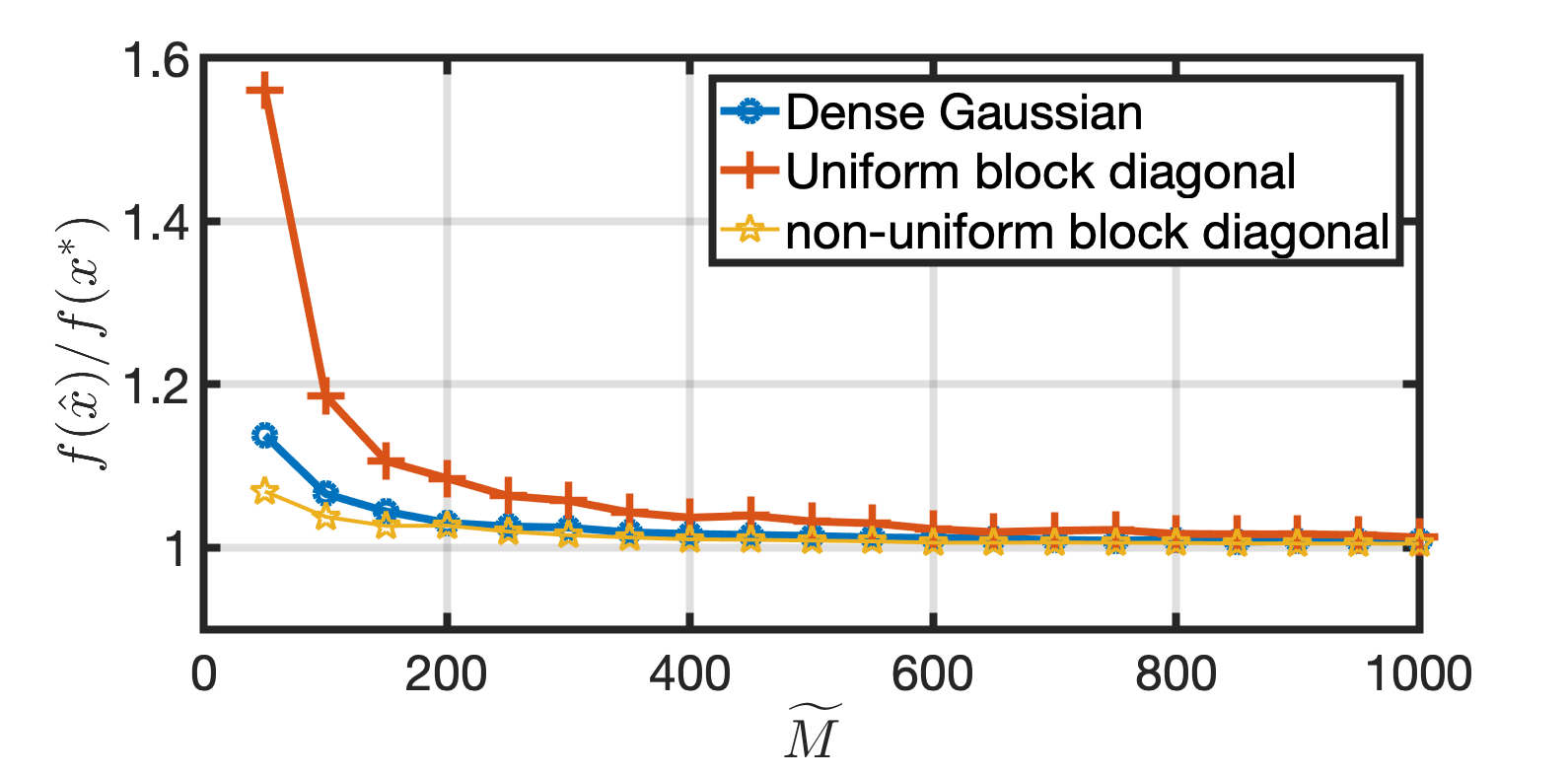}
    \caption{$f(\hat{x})/f(x^*)$ for three sketching matrices: a dense matrix with standard Gaussian entries, a block diagonal matrix with equal sized blocks (uniform diagonal matrix)  and a block diagonal matrix with entries designed as in \eqref{eq:strat} (non-uniform diagonal matrix). A ratio close to $1$ indicates that the sketching matrix is effective in solving \eqref{eq:sketch_ridge_reg_def}. When $\mj$'s are chosen appropriately, block diagonal matrices can be as effective as a general matrix.}
    \label{fig:rel_deltas_synth}
\end{figure}

We demonstrate the effectiveness of block diagonal sketching matrices by performing experiments on both synthetic and real data. In our first experiment, we demonstrate the importance of choosing the size of the diagonal blocks according to our proposed method given in \eqref{eq:strat}. We use the following parameters: $N = 2000$, $J = 10$, $d = 50$. We design the singular values such that for $\lambda = 0.15$, $\sd_\lambda = 8.5$, but $\rank(\mA) = 50$. For each trial, we generate $\mS$ with entries drawn from $\calN(0,1/\sqrt{\wm})$ and $\mS_D$ with the entries of $\mS_j$ drawn from $\calN(0,1/\sqrt{\mj})$. In Figure \ref{fig:rel_deltas_synth}, we plot $f(\hat{x})/f(x^*)$ averaged over $10$ trials for different values of $\wm$. In particular, we show that when $\mj = M_0 \Gamma(\mU_j)$, $\mS_D$ has the same rate of decay for $f(\hat{x})/f(x^*)$ as $\mS$, and has a worse rate otherwise. 

\begin{figure*}[t]
\includegraphics[scale = 0.15]{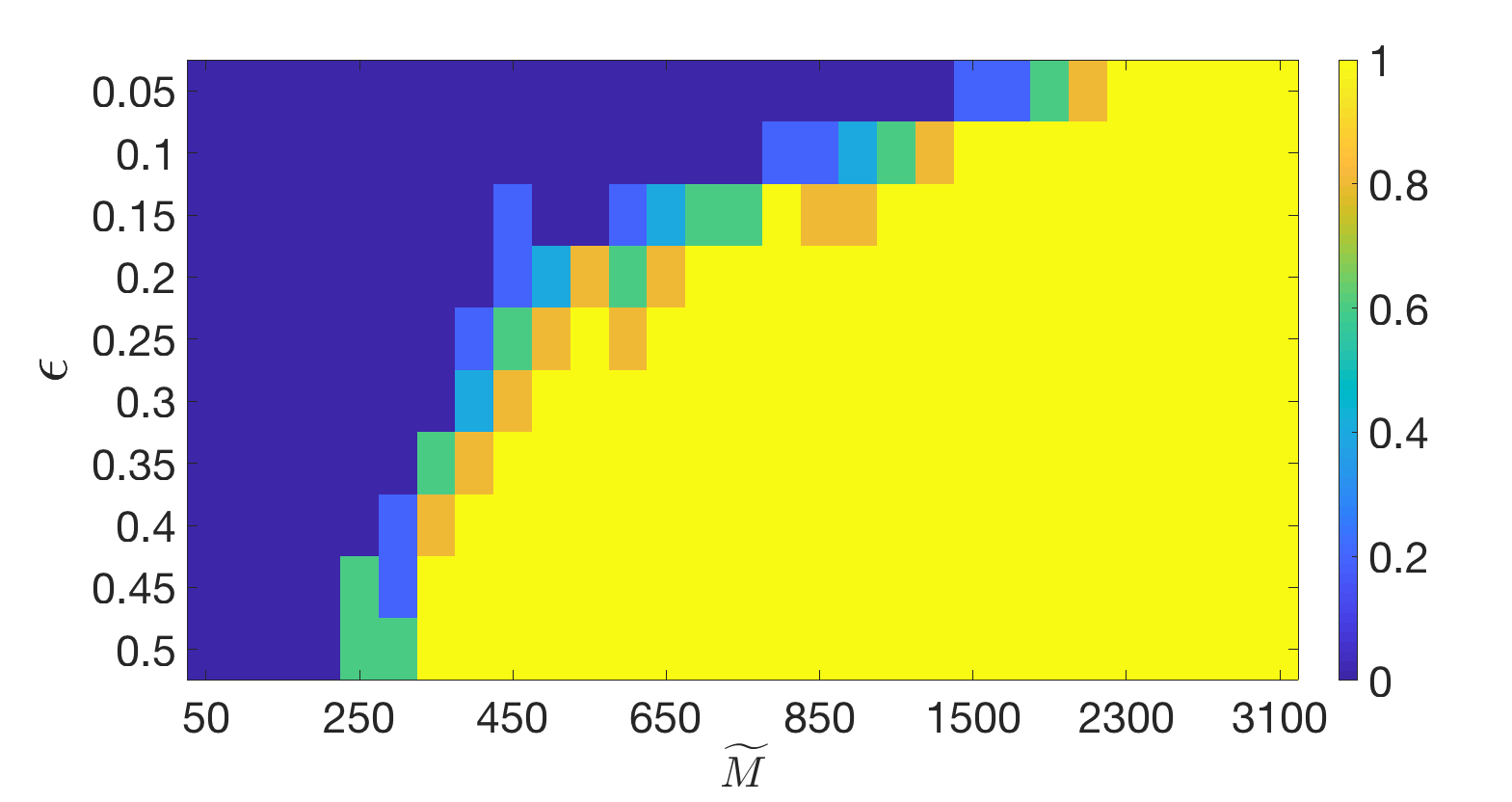}
\includegraphics[scale = 0.15]{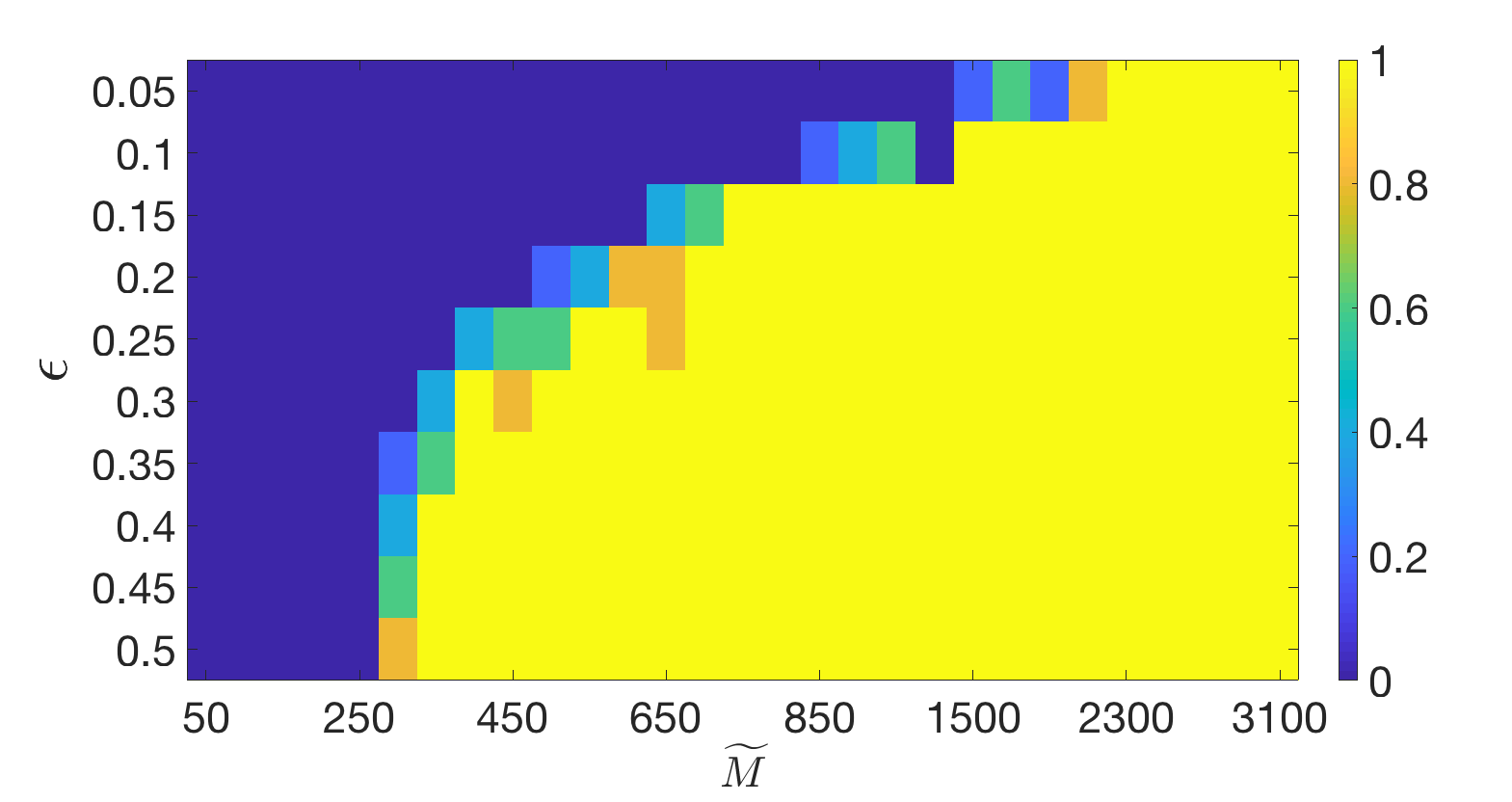}
\caption{Each plot shows the empirical probability of $f(\hat{x})\leq (1+\epsilon)f(x^*)$ for various values of $\wm$, computed using an average over 10 trials. The left pane is for results with dense matrices with sub-Gaussian entries, the right pane for results with block diagonal sketching matrices.}
\label{fig:real_data}
\end{figure*}

In our next set of experiments, we study performance in terms of prediction accuracy on the YearPredictionMSD dataset. It contains 89 audio features of a set of songs and the task is to predict their release year. The dataset has 463,715 training samples and 51,630 test samples. In this case, we use diagonal blocks of the same size. Across $10$ independent realizations of $\mS$ and $\mS_D$, we compute the empirical probability of $f(\hat{x})/f(x^*) \leq (1 + \epsilon)$ for various values of $\epsilon$ and $\wm$. We show phase transition plots in Figure \ref{fig:real_data} which demonstrate that block diagonal matrices are as effective as dense matrices in terms of accuracy, for the same sample complexity.

We also seek to highlight the computational advantages provided by block matrices. To this end, we compare the sketch compute times for block diagonal matrices with that of SRHT sketching matrices. We consider matrices $\mA$ of sizes $2^{18} \times 40$, $2^{20} \times 40$ and $2^{22} \times 40$ and divide them into $J = 2^{10}, \ 2^{12}, \ 2^{14}$ blocks respectively. In order to ensure fair comparison, we replace the SRHT matrix with randomly subsampled Fast Fourier transform (FFT) matrix, since both have the same theoretical sketch compute time, but the FFT matrix has very efficient software implementations. The sketch compute times are shown in Table \ref{table:runtime}. Our choice of $J$ renders each block small enough for very efficient computations. This results in block diagonal matrices being much faster compared to the FFT matrix.
\vspace{-2pt}
\section{Proof Sketch}
\label{sec:proof}
\vspace{-2pt}
In this section, we provide a sketch of the proof for both Theorems \ref{thm:DBD_apprx_mtx_prdct} and \ref{thm:DBD}. Full proofs are provided in the appendix. We first prove Theorem \ref{thm:DBD_apprx_mtx_prdct} and the proof for Theorem \ref{thm:DBD} follows by choosing $\mW$ and $\mY$ appropriately, as explained in Section \ref{subsec:bounds}. The fundamental property of a distribution of matrices $\calD$ that enables any $\mS \sim \calD$ to satisfy \eqref{eq:stbl_rnk_mtx_prduct} is the subspace embedding moment property, defined in \cite{sharper}:
\begin{equation}
    \label{eq:SE-moment}
    \underset{\mS \sim \calD}{\E}\norm{(\mS\mU)^T(\mS\mU) - \mId}^{l} \leq \epsilon^l \delta,
\end{equation} for some $l \geq 2$, where $\epsilon$ and $\delta$ are tolerance parameters that determine the sample complexity and $\mU$ is any orthobasis for the span of the columns of $\mW$ and $\mY$. Thus, our main goal is to prove the subspace embedding moment property holds for block diagonal sketching matrices.

Our methods differ from the common $\epsilon$-net argument, since using union bound for block diagonal matrices results in a suboptimal sample complexity. The main tools we use are the estimates for the suprema of chaos processes found in \cite{krahmerSOCP} and an entropy estimate from the study of restricted isometry properties of block diagonal matrices computed in \cite{RIPBD}. We first establish tail bounds on the spectral norm of the matrix 
\begin{equation}
    \label{eq:Delta}
    \mDelta =  (\mS_D\mU)^T(\mS_D\mU) - \mId,
\end{equation}
where $\mU$ is an orthobasis for a subspace of dimension $d$ and then bound its moments to establish the subspace embedding moment property. 

\subsection{Tail bound on the spectral norm of the matrix $\mDelta$}

We first express $\norm{\mDelta}$ as
\begin{align}
    \norm{\mDelta}  & = \sup_{\substack{\vz \in \R^d \\ \norm{\vz} = 1}} \left |\vz^T (\mS_D\mU)^T(\mS_D\mU)\vz - 1 \right |
    \nonumber \\
    & = \sup_{\substack{\vz \in \R^d \\ \norm{\vz} = 1}} \left |\norm{\mS_D\mU\vz}^2 - \E \norm{\mS_D\mU\vz}^2 \right |.
    \label{eq:sp_norm_reform}
\end{align}

For the matrices $\mS_j$, let $\vecd(\mS_j)$ denote their vectorized versions, obtained by stacking the columns one below the other. Let $\mS_v = [\vecd(\mS_1)^T \ \vecd(\mS_2)^T \ \cdots \vecd(\mS_J)^T]^T$ be the vector containing all of the $\vecd(\mS_j)$'s. Note that $\mS_v$ is a vector with entries drawn from $\calN(0,1)$. We can then express \eqref{eq:sp_norm_reform} as 
\begin{equation*}
    \norm{\mDelta} = \sup_{\mP_z \in \bfcalP} \left | \norm{\mP_z\mS_v}^2 - \E \norm{\mP_z\mS_v}^2 \right |
\end{equation*} where $\bfcalP$ is defined as 
\begin{align*}
    \bfcalP & = \left \{ \mP_z  =  \begin{bmatrix} \mP_1(z) & 0 & \cdots & 0 \\
    0 & \mP_2(z) & \cdots & 0 \\
    \vdots & \vdots & \ddots & \vdots \\
    0 & 0 & \cdots & \mP_J(z)
    \end{bmatrix}  \right \} \nonumber \\
    \mP_j(z) & = \frac{1}{\sqrt{\mj}} \begin{bmatrix}
    (U_1z)^T & 0 & \cdots & 0\\
    0 & (U_1z)^T & \cdots & 0 \\
    \vdots & \vdots & \ddots & \vdots \\
    0 & 0 & \cdots & (U_1z)^T
    \end{bmatrix}
\end{align*}
where $z \in \R^d$ and $\norm{z} = 1$. Observe that $\norm{\mDelta}$ is then the supremum of the deviation of a Gaussian quadratic form from its expectation, taken over the set $\bfcalP$. This matches the framework developed in \cite{krahmerSOCP} to bound such suprema. We use their result (Theorem 3.1, \cite{krahmerSOCP}) to obtain tail bounds on $\norm{\mDelta}$, stated in Lemma \ref{lm:delta_tail}.

\begin{lemma}
\label{lm:delta_tail}
For any orthonormal matrix $\mU \in R^{\wn \times d}$ and a block diagonal matrix $\mS_D$ as in Theorem \ref{thm:DBD_apprx_mtx_prdct}, there exists a constant $c$ such that
\begin{equation}
    \label{eq:tail_bound}
    \mathbb{P} \left  ( \norm{\mDelta} \leq c\sqrt{\frac{d\log (2/\delta) }{M_0}} \right ) \geq 1 - \delta.
\end{equation}
\end{lemma}

For a desired tolerance $\epsilon$, if $M_0 = \Omega\left ( \frac{d\log(2/\delta)}{\epsilon^2} \right )$, $\mathbb{P} \left  ( \norm{\mDelta} \leq \epsilon \right ) \geq 1 - \delta$. This is similar to a subspace embedding guarantee.  We now show that this tail bound naturally induces a bound on the moments of $\norm{\mDelta}$, from which the main theorems in section \ref{sec:main} can be proved.
\begin{table*}
\begin{center}
\renewcommand{\arraystretch}{2}
\begin{tabular}{ |c|c|c|c|c| } 
 \hline
 \multicolumn{5}{|c|}{Sketch compute time in seconds for large scale matrices} \\
 \hline
 $\wn, \ J$ & $\wm = 600$ & $\wm = 1400$ & $\wm =2200$ & $\wm = 3000$ \\
 \hline 
 $2^{18}, 2^{10}$ & $0.26 ; \mathbf{1.4\times 10^{-2}}$ & $0.26 ; \mathbf{2\times10^{-2}}$ & $0.26 ; \mathbf{3.88\cdot10^{-2}}$ & $0.26;\mathbf{4.2\times10^{-2}}$ \\
 \hline
 $2^{20}, 2^{12}$ & $1.16; \mathbf{2.7\times10^{-2}}$ & $1.16; \mathbf{3.9\times10^{-2}}$ & $1.16; \mathbf{5.1\times10^{-2}}$ & $1.16; \mathbf{6.3\times10^{-2}}$ \\ 
 \hline
 $2^{22}, 2^{14}$ & $5.87; \mathbf{7.9\times10^{-2}}$ & $5.87; \mathbf{9.1\times10^{-2}}$ & $5.87; \mathbf{11\times10^{-2}}$ & $5.86; \mathbf{11\times10^{-2}}$ \\
 \hline
\end{tabular}
\caption{Sketch compute time in sec. for various matrix sizes $\wn$ and sketch sizes $\wm$. In each cell, the left figure for FFT sketch and the right figure in boldface is for block diagonal matrices.}
\label{table:runtime}
\end{center}
\end{table*}

\subsection{Moment bound on $\norm{\mDelta}$}

Tail bounds for certain random variables can be translated into bounds on their moments using the following result:
\begin{lemma}[7.13,  \cite{foucart2013mathematical}]
\label{lm:tail}
Suppose that a random variable $q$ satisfies $
    \mathbb{P}\left ( |q| \geq \e^{1/\gamma} \alpha u \right ) \leq \beta \e^{-u^{\gamma}/\gamma}
$ for some $\gamma > 0$ and for all $u > 0$. Then, for $p > 0$, $
    \E |q|^p \leq  \beta \alpha^p (\e \gamma)^{p/\gamma} \Gamma \left ( \frac{p}{\gamma} + 1 \right )
$ where $\Gamma(\cdot)$ is the Gamma function. 
\end{lemma} 

By choosing $q = \norm{\mDelta}$, $\gamma = 2$, $\beta = 1$ and $\e^{-u^2/2} = \delta$, we obtain

\begin{lemma}
For any orthonormal matrix $\mU \in R^{\wn \times d}$ and a block diagonal matrix $\mS_D$ as in Theorem \ref{thm:DBD_apprx_mtx_prdct}, if  $M_0 = \Omega\left ( \frac{d\log(2/\delta)}{\epsilon^2} \right )$, then for $p = (\frac{\log(1/\delta)}{\epsilon^2})$, 
\begin{equation}
    \label{eq:moment}
    \E \norm{\mDelta}^p \leq \epsilon^p \delta
\end{equation} 
\end{lemma}
\textbf{Approximate matrix product guarantee}
 Let $\mW$ and $\mY$ be as in \eqref{eq:stbl_rnk_mtx_prduct}. As explained in \cite{optimal}, we can assume that they have orthogonal columns. For a given $k$ as in \eqref{eq:stbl_rnk_mtx_prduct}, let $\mW$ and $\mY$ be partitioned into groups of $k$ columns, with $\mW_l$ and $\mY_{l'}$ denoting the $l^{\text{th}}$ groups. The approach in \cite{optimal} then uses the following result in their argument, which follows from \eqref{eq:moment}: 
\begin{equation}
    \label{eq:argument}
    \E \norm{(\mS\mW_l)^T(\mS\mY_{l'}) - \mW_l^T\mY_{l'}}^p \leq \epsilon^p \norm{\mW_l}^p \norm{\mY_{l'}}^p \delta
\end{equation} for all pairs $(l,l')$. In their setting, this holds since the sketching is oblivious to the data matrices. Although block diagonal matrices are not oblivious, this result holds with for $M_0 = \Omega\left ( \frac{2k\log(2/\delta)}{\epsilon^2} \right )$. This is because of the observation that if $\mU$ is an orthobasis for the span of $\mW$ and $\mY$ and $\mU^{l,l'}$ is an orthobasis for the span of $\mW_l$ and $\mY_{l'}$ , then 
$   \Gamma(\mU^{l,l'}_j) \leq \Gamma(\mU_j) $ for all pairs $(l,l')$. Hence, a given block diagonal sketching matrix $\mS_D$ can satisfy \eqref{eq:argument}. The rest of the proof remains the same as in \cite{optimal}. This concludes the proof for Theorem \ref{thm:DBD_apprx_mtx_prdct}.

\section{Conclusion}

In this paper, we study a particular model that can be used while applying sketching techniques to high dimensional data that are available in a distributed fashion. Our proposed block diagonal sketching model forms an intermediate model between sampling methods and random projection methods and is a useful abstraction. We show theoretically and experimentally that choosing the sketch sizes proportional to a certain coherence term of the data blocks results in an optimal sample complexity. While we do not provide formal analysis of the algorithm to estimate the coherence parameters, we show empirically that they can be estimated. 

\appendix 

\section{Proof of Theorem 1}

 The fundamental property of a distribution of matrices $\calD$ that enables any $\mS \sim \calD$ to satisfy (8, main paper) is the subspace embedding moment property, defined in \cite{sharper}:
\begin{equation}
    \label{eq:SE-moment}
    \underset{\mS \sim \calD}{\E}\norm{(\mS\mU)^T(\mS\mU) - \mId}^{l} \leq \epsilon^l \delta,
\end{equation} for some $l \geq 2$, where $\epsilon$ and $\delta$ are tolerance parameters that determine the sample complexity and $\mU$ is any orthobasis for the span of the columns of $\mW$ and $\mY$. Thus, our main goal is to prove the subspace embedding moment property holds for block diagonal sketching matrices.

Our methods differ from the common $\epsilon$-net argument, since using union bound for block diagonal matrices results in a suboptimal sample complexity. The main tools we use are the estimates for the suprema of chaos processes found in \cite{krahmerSOCP} and an entropy estimate from the study of restricted isometry properties of block diagonal matrices computed in \cite{RIPBD}. We first establish tail bounds on the spectral norm of the matrix 
\begin{equation}
    \label{eq:Delta}
    \mDelta =  (\mS_D\mU)^T(\mS_D\mU) - \mId,
\end{equation}
where $\mU$ is an orthobasis for a subspace of dimension $d$ and then bound its moments to establish the subspace embedding moment property.

\subsection{Suprema of chaos processes}

We briefly state here the main result from \cite{krahmerSOCP} that provides a uniform bound on the deviation of a Gaussian quadratic form from its expectation. Obtaining a tail bound on the spectral norm of $\mDelta$ is just a particular application of this general framework.

For a given set of matrices $\bfcalP$, we define the spectral radius $d_2(\bfcalP)$, the Frobenius norm radius $d_F(\bfcalP)$, and the Talagrand functional $\gamma_2(\bfcalP, \|\cdot\|_2)$ as 
\begin{align}
    d_2(\bfcalP) & = \sup_{\mP \in \bfcalP} \norm{\mP}, \nonumber \\
    d_F(\bfcalP) & = \sup_{\mP \in \bfcalP} \norm{\mP}_F, \nonumber \\
    \gamma_2(\bfcalP, \|\cdot\|_2) & = \int_0^{d_2(\bfcalP)} \sqrt{\log N (\bfcalP, \| \|_2, u)} du ,\nonumber
\end{align}  where $N (\bfcalP, \| \|_2, u)$ denotes the covering number of the set $\bfcalP$ with respect to balls of radius $u$ in the spectral norm. The main result of \cite{krahmerSOCP} then is the following theorem. 

\begin{theorem}
\label{thm:SupremaChaos}
[Theorem 3.1, \cite{krahmerSOCP} ] Let $\bfcalP$ be a set of matrices and let $\phi$ be a vector of i.i.d. standard normal entries. Then for $t \geq 0$, 
\begin{equation}
\mathbb{P} \left( \sup_{\mP \in \bfcalP}|\|\mP\phi\|^2 - \E\| \mP\phi\|^2 | > c_1E+t \right) \leq 2 \e^{-c_2\min\{\frac{t^2}{V^2} , \frac{t}{U}\}}
\end{equation}
where 
\begin{align*}
    E &=  \gamma_2(\bfcalP)[ \gamma_2(\bfcalP) +d_F(\bfcalP) ] + d_2(\bfcalP)d_F(\bfcalP), \\
    V &=  d_2(\bfcalP)[ \gamma_2(\bfcalP) + d_F(\bfcalP)], \\
    U &= d_2^2(\bfcalP).
\end{align*}
\end{theorem} 
A similar approach of using the results from \cite{krahmerSOCP} to analyze block diagonal random matrices was first used in \cite{RIPBD} in the context of compressed sensing. However, we target a different set of problems that result in different theoretical considerations and proof techniques.

\subsection{Tail bound on the spectral norm of the matrix $\mDelta$}

We first express $\norm{\mDelta}$ as
\begin{align}
    \norm{\mDelta} & = \sup_{\substack{\vz \in \R^d \\ \norm{\vz} = 1}} \left |\vz^T (\mS_D\mU)^T(\mS_D\mU)\vz - 1 \right | 
    \label{eq:sp_norm_reform}\\
    & = \sup_{\substack{\vz \in \R^d \\ \norm{\vz} = 1}} \left |\norm{\mS_D\mU\vz}^2 - \E \norm{\mS_D\mU\vz}^2 \right |.
\end{align}
For the matrices $\mS_j$, let $\vecd(\mS_j)$ denote their vectorized versions, obtained by stacking the columns one below the other. Let $\mS_v = [\vecd(\mS_1)^T \ \vecd(\mS_2)^T \ \cdots \vecd(\mS_J)^T]^T$ be the vector containing all of the $\vecd(\mS_j)$'s. Note that $\mS_v$ is a vector with entries drawn from $\calN(0,1)$. We can then express \eqref{eq:sp_norm_reform} as 
\begin{equation*}
    \norm{\mDelta} = \sup_{\mP_z \in \bfcalP} \left | \norm{\mP_z\mS_v}^2 - \E \norm{\mP_z\mS_v}^2 \right |
\end{equation*} where $\bfcalP$ is defined 
\begin{align*}
    \bfcalP &= \left \{ \mP_z  =  \begin{bmatrix} \mP_1(z) & 0 & \cdots & 0 \\
    0 & \mP_2(z) & \cdots & 0 \\
    \vdots & \vdots & \ddots & \vdots \\
    0 & 0 & \cdots & \mP_J(z)
    \end{bmatrix}  \right \}, \\ \mP_j(z) &= \frac{1}{\sqrt{\mj}} \begin{bmatrix}
    (U_1z)^T & 0 & \cdots & 0\\
    0 & (U_1z)^T & \cdots & 0 \\
    \vdots & \vdots & \ddots & \vdots \\
    0 & 0 & \cdots & (U_1z)^T
    \end{bmatrix}
\end{align*}
where $z \in \R^d$ and $\norm{z} = 1$. Observe that $\norm{\mDelta}$ is then the supremum of the deviation of a Gaussian quadratic form from its expectation, taken over the set $\bfcalP$. 

We can then compute the corresponding quantities $d_2(\bfcalP)$, $d_F(\bfcalP)$ and $\gamma_2(\bfcalP, \norm{\cdot}_2)$ as follows.  

The spectral radius $d_2(\calP)$ is defined as 

\begin{align*}
    \label{eq:d2}
    \sup_{\mP_z \in \calP} \norm{\mP_z} & = \max_{j,\norm{z}_2 = 1} \frac{\norm{U_jz}}{\sqrt{\mj}}\\
    & \leq \min \left( \frac{\sqrt{N}\norm{
    U_j}_\infty \norm{z}_1}{\sqrt{\mj}} , \frac{\norm{U_j}\norm{z}_2}{\sqrt{\mj}}\right ) \\
    & \leq \min \left( \frac{\sqrt{N}\norm{
    U_j}_\infty \norm{z}_1}{\sqrt{\mj}} , \frac{\norm{U_j}\norm{z}_1}{\sqrt{\mj}}\right )\\
    & \leq \norm{z}_1 /\sqrt{M_0} \leq  \frac{d_1}{\sqrt{M_0}}
\end{align*}
where the fourth line follows from the definition of $\mj$.

The radius in the Frobenius norm $d_F(\calP)$ is defined as 

\begin{align*}
    \sup_{\mP_z \in \calP} \norm{\mP_z}_F & = \sum_j {\norm{U_jz}^2} = 1.
\end{align*}

The upper bound for $\gamma_2(\calP, \norm{\cdot})$ can be obtained from the Equation (34) in Eftekhari et al., 2015).. In their derivation, they consider a full orthobasis and the set of $d$-sparse vectors. This bound also holds for a fixed $d$-dimensional subspace. Hence, 

\begin{equation}
    \gamma_2(\calP, \norm{\cdot}) \lesssim \sqrt{\frac{d}{M_0}} \log d \log \widetilde{M}
\end{equation}

Plugging these quantities into Theorem \ref{thm:SupremaChaos}, we can obtain Lemma 1. 

\begin{lemma}
\label{lm:delta_tail}
For any orthonormal matrix $\mU \in R^{\wn \times d}$ and a block diagonal matrix $\mS_D$ as in Theorem 1, there exists a constant $c$ such that
\begin{equation}
    \label{eq:tail_bound}
    \mathbb{P} \left  ( \norm{\mDelta} \leq c\sqrt{\frac{d\log (2/\delta) }{M_0}} \right ) \geq 1 - \delta.
\end{equation}
\end{lemma}

For a desired tolerance $\epsilon$, if $M_0 = \Omega\left ( \frac{d\log(2/\delta)}{\epsilon^2} \right )$, $\mathbb{P} \left  ( \norm{\mDelta} \leq \epsilon \right ) \geq 1 - \delta$. This is similar to a subspace embedding guarantee.  We now show that this tail bound naturally induces a bound on the moments of $\norm{\mDelta}$, from which the main theorems in Section 2 can be proved.

\subsection{Moment bound on $\norm{\mDelta}$}

Tail bounds for certain random variables can be translated into bounds on their moments using the following result:
\begin{lemma}[Proposition 7.13, \cite{foucart2013mathematical}]
\label{lm:tail}
Suppose that a random variable $q$ satisfies, for some $\gamma > 0$, \begin{equation*}
    \mathbb{P}\left ( |q| \geq \e^{1/\gamma} \alpha u \right ) \leq \beta \e^{-u^{\gamma}/\gamma}
\end{equation*} for all $u > 0$. Then, for $p > 0$,
\begin{equation*}
    \E |q|^p \leq  \beta \alpha^p (\e \gamma)^{p/\gamma} \Gamma \left ( \frac{p}{\gamma} + 1 \right )
\end{equation*}
where $\Gamma(\cdot)$ is the Gamma function. 
\end{lemma} 

To adapt this result to bound the moments of the spectral norm of the random matrix $\mDelta$, we can choose $q = \norm{\mDelta}$, $\gamma = 2$, $\beta = 1$ and $\e^{-u^2/2} = \delta$. We can then obtain the following result.

\begin{lemma}
For any orthonormal matrix $\mU \in R^{\wn \times d}$ and a block diagonal matrix $\mS_D$ as in Theorem 1 and  $M_0 = \Omega\left ( \frac{d\log(2/\delta)}{\epsilon^2} \right )$, then 
\begin{equation}
    \label{eq:moment}
    \E \norm{\mDelta}^p \leq \epsilon^p \delta
\end{equation} for $p = (\frac{\log(1/\delta)}{\epsilon^2})$.
\end{lemma}

\subsection{Approximate matrix product guarantee}

With the moment bound established above, we can now use the framework given by \cite{optimal} to establish (8, main paper). However, we cannot use their proof directly, since the sample complexity $\wm$ in the moment bound in \eqref{eq:moment} is not oblivious to the matrix $\mU$. However, once we fix the data matrix, we can adapt the argument used in \cite{optimal} to show that (8, main paper) holds. 

Let $\mW$ and $\mY$ be as in (8, main paper). As explained in \cite{optimal}, we can assume that they have orthogonal columns. For a given $k$ as in (8, main paper), let $\mW$ and $\mY$ be partitioned into groups of $k$ columns, with $\mW_l$ and $\mY_{l'}$ denoting the $l^{\text{th}}$ groups. \cite{optimal} then use the following result in their argument, which follows from \eqref{eq:moment}: 
\begin{equation}
    \label{eq:argument}
    \E \norm{(\mS\mW_l)^T(\mS\mY_{l'}) - \mW_l^T\mY_{l'}}^p \leq \epsilon^p \norm{\mW_l}^p \norm{\mY_{l'}}^p \delta
\end{equation} for all pairs $(l,l')$. This holds since in their setting, the sketching matrices are oblivious to the data matrices. 

Although block diagonal matrices are not oblivious, this result holds with for $M_0 = \Omega\left ( \frac{2k\log(2/\delta)}{\epsilon^2} \right )$. This is because of the observation that if $\mU$ is an orthobasis for the span of $\mW$ and $\mY$ and $\mU^{l,l'}$ is an orthobasis for the span of $\mW_l$ and $\mY_{l'}$ , then 
\begin{equation}
    \label{eq:blockGamma}
    \Gamma(\mU^{l,l'}_j) \leq \Gamma(\mU_j)
\end{equation} for all pairs $(l,l')$. Hence, a given block diagonal sketching matrix $\mS_D$ can satisfy \eqref{eq:argument} as well. The rest of the proof remains the same as \cite{optimal}. This concludes the proof for Theorem 1. Extending this to prove Theorem 2 is straightforward, with $\mS_D$ being a particular case of their framework.

\section{Algorithm for estimation of the incoherence parameters $\Gamma(\mU_j)$}

Our algorithm for estimating the block incoherence parameters is inspired by the algorithms for leverage score estimation in the row sampling literature \cite{drineas2012fast,Woodruff:2014:STN:2693651.2693652} and from randomized SVD algorithms \cite{halko2011finding}.

The main idea is the following: suppose we had access to the QR factorization of the data matrix $\mA \in \wn \times d$:
\begin{equation}
    \mA = \mQ \mR.
\end{equation}Then, an orthobasis can be obtained by computing $\mQ = \mA \mR^{-1}$. However, computing the QR-factorization is as expensive as the matrix multiplication or ridge regression problems. We use a similar approach, but we only aim to capture the row space of $\mA$ in a distributed fashion. However, we take random projections in an iterative fashion, until the row space of the sketch ``converges". we estimate the QR factorization from this resulting sketch. 
 Our algorithm is described in Algorithm \ref{algo_gamma}. Note that we only aim to compute a constant factor approximation of the QR factors. Hence, computing the $\mR$ takes, in the worst case, $O(J d N\log N) = O(\wn d \log N)$ time. The QR factorization in each iteration can be updated from its previous estimates efficiently. Computing the final estimate takes about $O(Jd^3)$ time. Finally computing $\Hat{\Gamma}(\mU_j)$'s takes $O(\wn d)$ time, resulting in a total worst case time complexity of $O(\wn d \log N)$.

%
%
%
%

\begin{algorithm}
\SetAlgoLined
\KwResult{Normalized estimates $\Hat{\Gamma}(\mU_j)/\sum_j \Hat{\Gamma}(\mU_j) $}
 Initialize $\Omega \in \R^{O(1) \times N} , \mQ = 0, \mR = 0$, $\Hat{\mA} = 0$  where $\Omega$ is drawn from any subsampled FJLT. \\
 \While{rank(R) not converged}{
  Compute $\hat{\mA}_j = \Omega \mA $ \;
    Aggregate $\Hat{\mA} = [ \hat{\mA}_1^\top \ \hat{\mA}_2^\top  \cdots \hat{\mA}_J^\top]^\top$ at the central processing unit with previous estimate  \;
    Update $\mQ \mR$ = qr($\Hat{\mA}$)  \;
    Draw a new independent realization of $\Omega$ \;
  
 }
 Compute $\Hat{\Gamma}(\mU_j) = \norm{\mA_j\mR^{-1}}_F^2$ \;
 \caption{Estimation of incoherence parameters up to constant factor error}
\end{algorithm}

\bibliographystyle{unsrt}
\bibliography{bibliography}

\end{document}